\documentclass[10pt,twocolumn,letterpaper]{article}

\usepackage{cvpr}
\usepackage{times}
\usepackage{epsfig}
\usepackage{graphicx}
\usepackage{amsmath}
\usepackage{amssymb}
\usepackage{soul}
\usepackage[usenames, dvipsnames]{color}

\cvprfinalcopy 

\def\cvprPaperID{****} 

\begin{document}

\title{Video Object Segmentation using Tracked Object Proposals}

\author{Gilad Sharir
		\and
		Eddie Smolyansky\\
		Visualead\\
		Hertzeliya, Israel\\
		{\tt\small \{gilad,eddie,itamar\}@visualead.com}
		\and 
		Itamar Friedman
}

\maketitle

\begin{abstract}
We present an approach to semi-supervised video object segmentation, in the context of the DAVIS 2017~\cite{Pont-Tuset_arXiv_2017} challenge. Our approach combines category-based object detection, category-independent object appearance segmentation and temporal object tracking.
We are motivated by the fact that the object’s semantic category tends not to change throughout the video while its appearance and location can vary considerably. In order to capture the specific object appearance independent of its category, for each video we train a fully convolutional network using augmentations of the given annotated frame. We refine the appearance segmentation mask with the bounding boxes provided either by a semantic object detection network, when applicable, or by a previous frame prediction. By introducing a temporal continuity constraint on the detected boxes, we are able to improve the object segmentation mask of the appearance network, and achieve competitive results on the DAVIS datasets. 
\end{abstract}


\section{Introduction}
Video object segmentation is a key problem in computer vision, and has many useful applications in the real world. Several such applications include: content retrieval from large video databases, video editing, and monitoring of surveillance videos. In the context of object segmentation there are two main problem scenarios: semi-supervised and unsupervised. In the former, a single annotated frame is given to the system as input along with the video, while in the latter, only the video frames are given. In both cases the objective is to output a segmentation mask that separates the main object from the background. Our method applies to the semi-supervised scenario. 

A method named OSVOS~\cite{Cae+17} that was recently proposed to solve this problem trains an appearance network using the annotated frame as a one-shot training example and applies this network to the rest of the video frames to infer their segmentation mask. In principle, the segmentation obtained from the one-shot network has two major weaknesses. The first is when the video contains more than one instance of an object similar to the annotated one. In many such cases the appearance network mistakenly recognizes all or several such instances as part of the object. The second is when an object's appearance changes drastically during the video, and the object in the first frame bears little resemblance to its appearance in future frames. In such cases the appearance network recognizes the object well in the first frames of the video, but the segmentation quickly deteriorates as the video progresses.

Our proposed method leverages the annotated frame as training data to an appearance network in the same manner as in~\cite{Cae+17}, and additionally uses semantic object detection in order to alleviate the aforementioned drawbacks. Our method uses the appearance net and temporal-constraints to filter the bounding boxes proposed by the object-detector, and in turn the selected bounding boxes help us filter the correct connected components from the segmentation mask. In the following sections we will explain this method in more detail. 

Additionally, we have built a full demonstration mobile app to test and evaluate our general object segmentation capability in the real world, including a spatio-temporal interactive representation of an object and user-assisted first frame segmentation. This has allowed us to collect our own internal dataset (soon to be published), which yields similar test results to DAVIS-2016.

The contribution of this paper can be summarized as follows: 
i. We suggest a way of utilizing semantic object detection (bounding boxes) to improve the quality of general object segmentation. 
ii. We suggest a mechanism to leverage temporal information by tracking objects across frames, both for uncategorized objects and objects belonging to a semantic class.
iii. We suggest a simplification to the basic appearance net (OSVOS) which enables an order of magnitude faster training with comparable tests results.
\newpage
\section{Related Work}
{\bf Video object segmentation} There are several methods that attempt to solve the problem of category independent object segmentation in videos, in a semi-supervised approach. One work that greatly inspired us is OSVOS~\cite{Cae+17}. 
The main difference between our approach and theirs is that we additionally use bounding box detection to define focus areas on top of the computation of the category-independent object foreground segmentation. 

Another similar approach is that of SGV~\cite{sgv-cae}, which employs a semantic (category-dependent) segmentation map and wisely apply a conditional classifier to improve the foreground segmentation. We found that rather than computing semantic segmentation maps, it is possible to use semantic or general bounding box detections to focus on the annotated objects, followed by processing of connected components in the focus areas to achieve similar results. In addition, we force temporal coherency on the bounding boxes by tracking them across frames.


Another approach~\cite{Perazzi2017} applies a model to each frame and uses the segmentation mask of each frame as input to the model for segmenting the next frame. This is similar to our approach in that it uses temporal continuity between frames to localize the object across the video. 

{\bf Instance segmentation} A closely related field is that of instance segmentation, in which the goal is to provide segmentation masks of each object instance. Several works attempt to solve this task using multi-task network cascades~\cite{mnc-dai}, or fully convolutional networks~\cite{dai-eccv16}. Our method differs in that our goal is to segment an object instance for a whole video, while the object category is irrelevant.

{\bf Object detection} Object detection is an important area of research in computer vision. One of the leading methods is Faster-RCNN~\cite{faster-nips}. Although this method performs category-dependent detection, we are able to use in cases where an object gets a high confidence semantic detection. Along with a temporal tracking method, we show how we can enforce coherent segmentation throughout the video.  

\section{Our Architecture}

To facilitate easier reading of the rest of the paper, let us propose some terminology:
"map" refers to a grayscale image result, while "mask" refers to a binary image.
"Segmentation" refers to a pixel-level result, while "detection" refers to a bounding box.
"Semantic" refers to  a category-based result (i.e. the result belongs to one of X categories).


\begin{figure*}
\centering
\includegraphics[width=\linewidth]{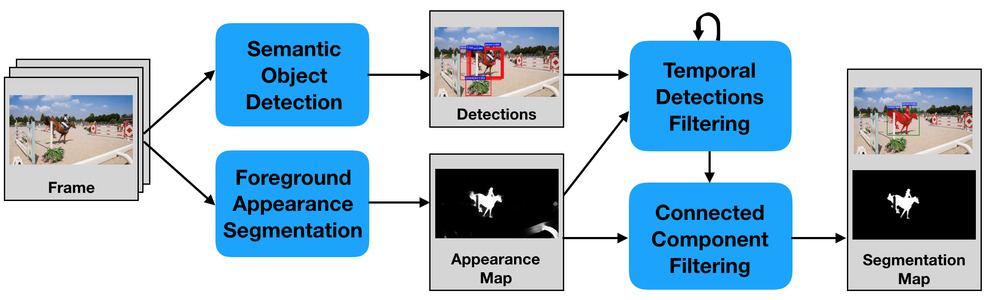}
\caption{Overview of our system.}
\label{figure:system}
\end{figure*}

\subsection{Appearance Network} \label{Appearance Network}
First, each input frame is passed through the appearance net from which category-independent object segmentations are obtained. Following previous work {~\cite{Cae+17}}, this net is based on the VGG16 architecture and has been transformed to a fully convolutional network. Unlike FCN~\cite{Long_2015_FCN} and in order to preserve spatial resolution, the final pooling layer and the fully connected layers have been stripped completely. 

Skip connections are employed to allow multi-resolution spatial information to flow from the shallow layers to the end of the net and improve the segmentation accuracy over the fine details of the object's contour. More specifically, we take the final feature map of each stage of VGG16 before the pooling layer and convolve it with a single $1X1$ kernel to get an output of depth $1$ (a grayscale segmentation probability map of the same size as current down-sampled stage) and up-sample it with a bilinear filter to the original image size.

Finally, at the end of the net these side outputs are concatenated and passed through a fusion convolutional layer that outputs the net prediction: a full sized, grayscale segmentation probability map. 
To achieve pixel-level segmentation, the softmax classifier is replaced by a class-balanced sigmoid cross-entropy loss layer that provide a binary classification mask (as described in~\cite{xie2015HED}).

We train the appearance net on the frames of the training videos in the DAVIS dataset, and learn a general object appearance network. Borrowing the terms from~\cite{Cae+17}, this is called the {\bf parent network}. We later apply this parent network to each test video separately by further training it on the single annotated frame provided for each video (one-shot training). Thus, at inference time we fine-tune the network to focus on the specific object of interest.

\subsection{Object Detection Network}
Concurrently, the frame is passed through an instance level semantic object detection network. This net takes the original RGB image as input and produces a set of bounding boxes for any object that it finds belonging to a collection of classes it supports. Our motivation is that many objects in natural videos belong to one semantic class or another (and those that don't are still often classified into the semantic set; For example, a camel may be correctly detected but misclassified as a horse). 

This net's main contribution is that it separates instances of the same object class, thus allowing enabling the choice of the correct instance in videos where more than one of the same class has been selected by the appearance net.
The specific algorithm or implementation used for semantic object detection is transparent to us, as long as it works on an instance level. Out of convenience, we have chosen to work with an implementation of Faster R-CNN based on the ResNet-101 architecture.

\subsection{Bounding Box Filtering}
\subsubsection{Appearance Based Filter}

After passing the input frame through both networks, we have an initial segmentation prediction map obtained from the one-shot appearance network and also many bounding box proposals for the objects that the semantic detection network has identified. We now describe a method for combining the results of the two networks to refine a final prediction object segmentation map for each frame in the video. First, we select the bounding boxes belonging to our annotated objects using the ground truth of the first image. Then we keep choosing the right bounding boxes in future frames by searching for bounding box proposals that best match the appearance map and by forcing temporal continuity on these detections.

For the first image, we choose the semantic detections (bounding boxes) that have the best overlap with the object segmentation given by the first frame ground truth. The selected classes are then stored in memory, to be searched for in later frames.

For any frame thereafter, only classes that have been found in the first frame are of interest and the rest are discarded. Out of the remaining detection object proposals we choose those that best fit the appearance map prediction according to a measure of intersection over union between each bounding box proposal and the appearance map.

\begin{figure*}
\centering
\includegraphics[width=0.24\textwidth]{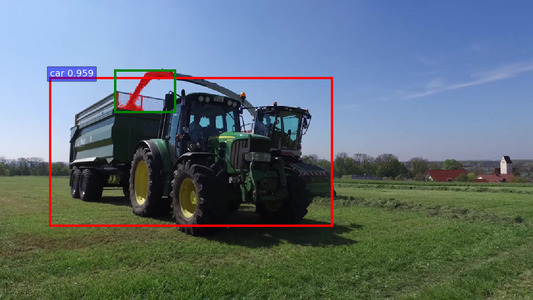}
\includegraphics[width=0.24\textwidth]{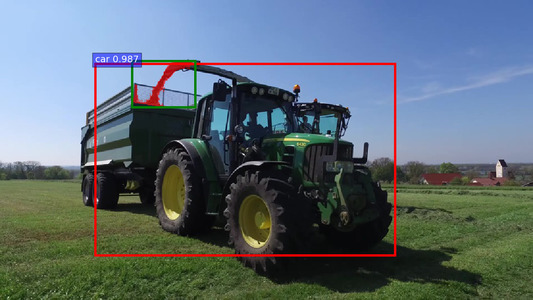} 
\includegraphics[width=0.24\textwidth]{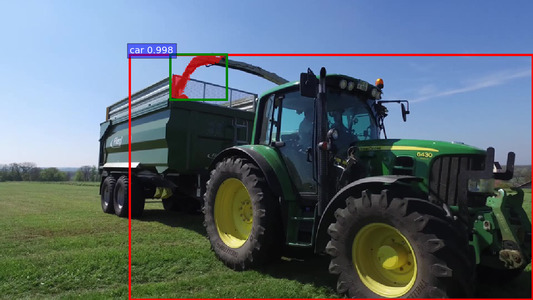}
\includegraphics[width=0.24\textwidth]{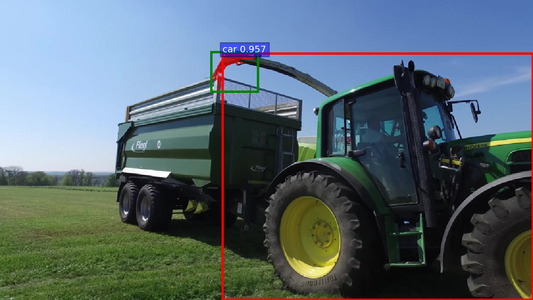} \\
\vspace{0.05cm}
\includegraphics[width=0.24\textwidth]{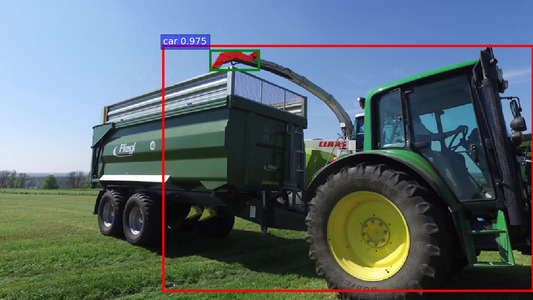}
\includegraphics[width=0.24\textwidth]{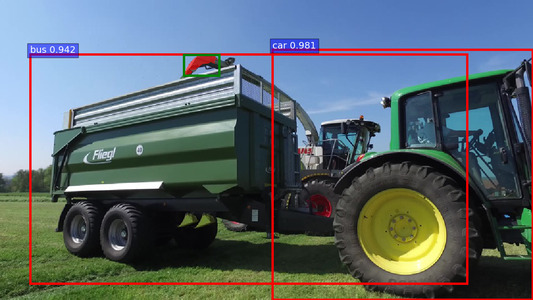} 
\includegraphics[width=0.24\textwidth]{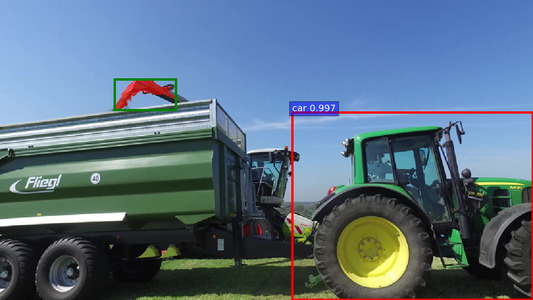}
\includegraphics[width=0.24\textwidth]{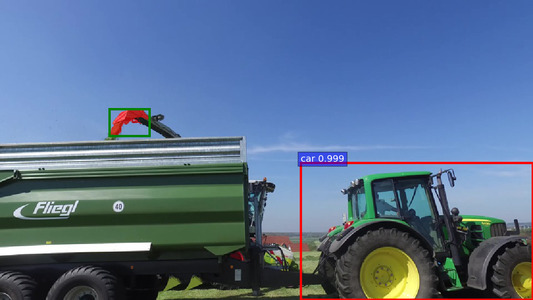}
\caption{An uncategorized object tracking example. Note the temporal tracking on the non-semantic mowed grass object.}
\label{fig:uncategorized}
\end{figure*}

\subsubsection{Temporal Filter} \label{temporal}

\begin{figure*}
\centering
\includegraphics[width=0.24\textwidth]{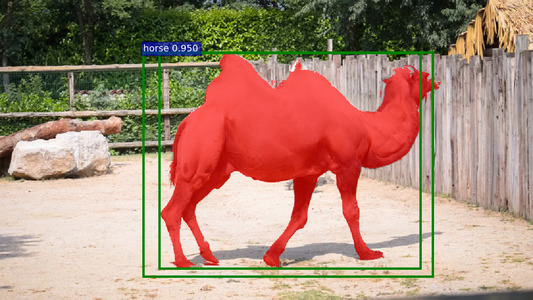}
\includegraphics[width=0.24\textwidth]{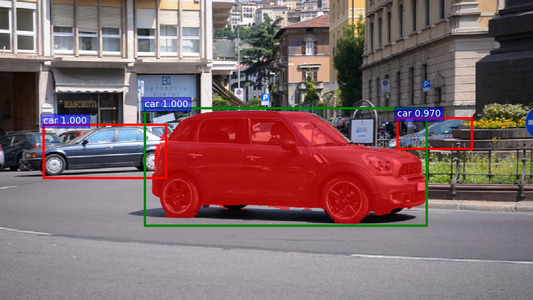}
\includegraphics[width=0.24\textwidth]{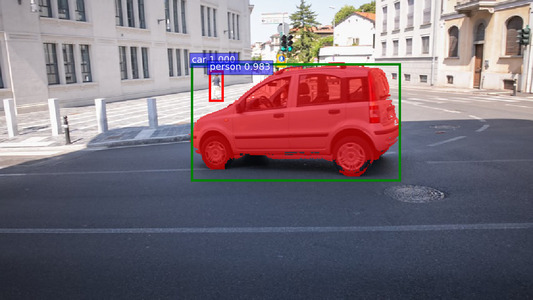}
\includegraphics[width=0.24\textwidth]{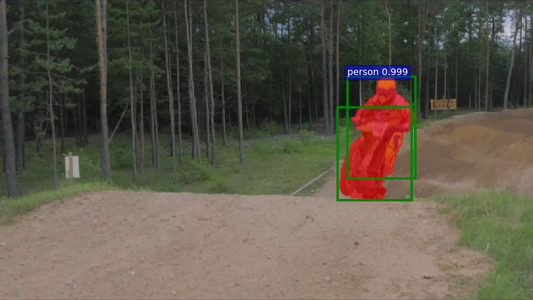}\\
\vspace{0.1cm}
\includegraphics[width=0.24\textwidth]{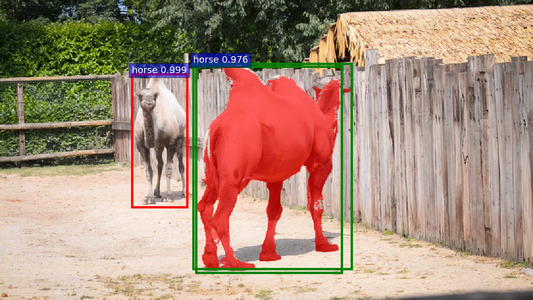} 
\includegraphics[width=0.24\textwidth]{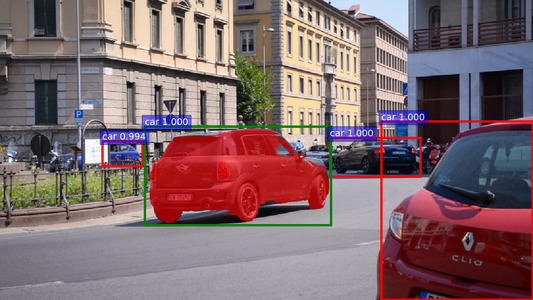}
\includegraphics[width=0.24\textwidth]{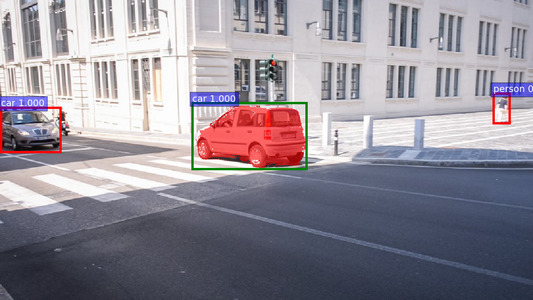}
\includegraphics[width=0.24\textwidth]{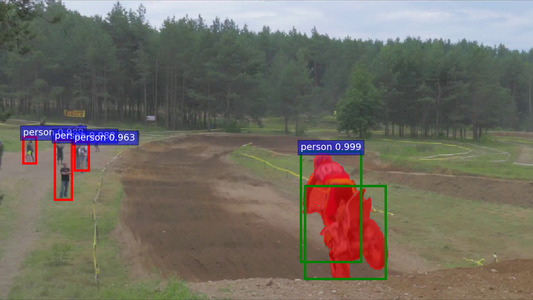}\\
    
   \caption{Bounding box tracking. {\bf First row}: First frame in the sequence. {\bf Second row}: Successful localization of the original object instance (in green) after a number of frames by tracking the bounding box}.
\label{fig:short}
\end{figure*}

Having selected the correct bounding box of a semantic object in the previous frame, we now run the risk of switching to another object instance whose appearance prediction highly overlaps with its semantic bounding box. To further ensure a correct selection of the bounding boxes, we now perform temporal tracking on the correct bounding box by filtering only those boxes that pass an intersection over union threshold with the object location in the previous frame (see figure~\ref{fig:short}). 

If the semantic object detection fails to detect any object in the first frame (due to an unrecognized category, see example in figure~\ref{fig:uncategorized}), we
instead use the first frame annotation to define a bounding box. Then, for all future frames we find connected components that intersect with the previous bounding box and remove all other segments, and finally choose a new bounding box according to the connected components that were chosen








By the end of this stage we should have an appearance map and the correct semantic bounding box detections of our annotated objects.


\begin{figure*}
\centering
\includegraphics[width=0.24\textwidth]{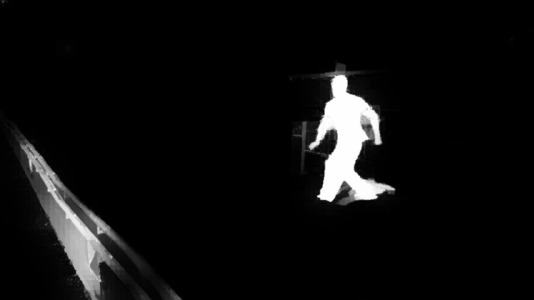}
\includegraphics[width=0.24\textwidth]{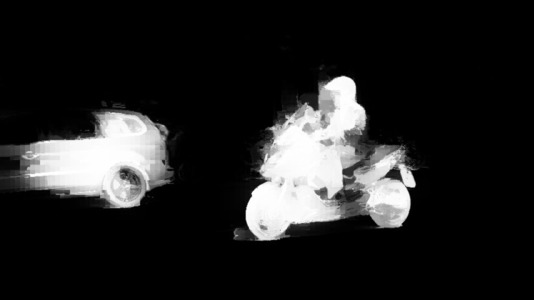} 
\includegraphics[width=0.24\textwidth]{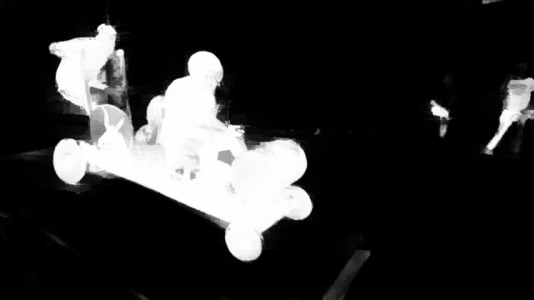} 
\includegraphics[width=0.24\textwidth]{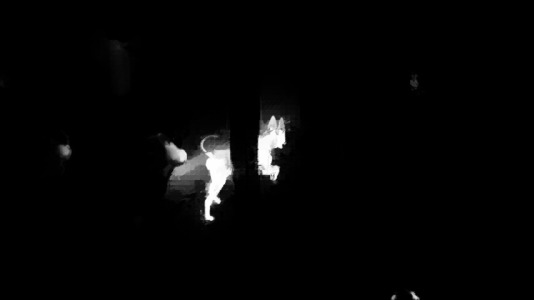}\\
\vspace{0.1cm}
\includegraphics[width=0.24\textwidth]{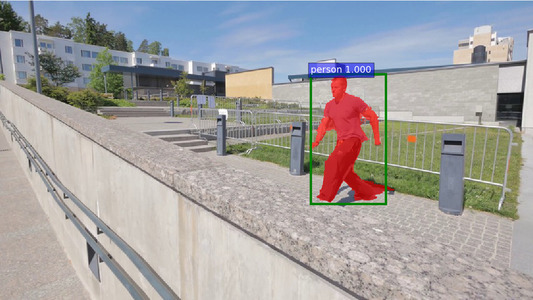} 
\includegraphics[width=0.24\textwidth]{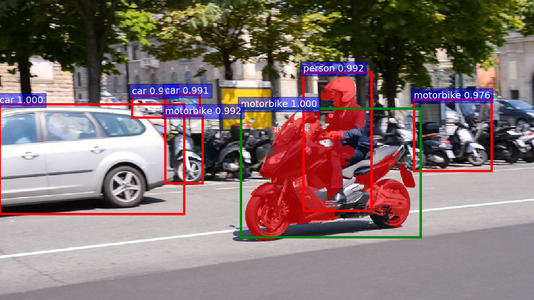} 
\includegraphics[width=0.24\textwidth]{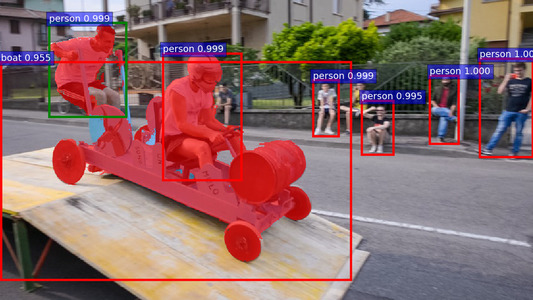}
\includegraphics[width=0.24\textwidth]{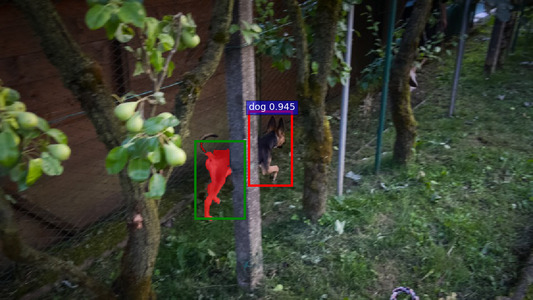}\\
 \medskip
\caption{{\bf Top}: The output of the appearance network. {\bf Bottom}: Final result after using the detected bounding box to filter the appearance map. 
The right column shows a failure case: the occluded dog is recognized as two separate instances by the detection network and since the instances are not connected, one of them was removed from the final segmentation map.}
\label{figure:filter}
\end{figure*}

\subsection{Connected Components Filter} \label{conncomp}
In the final stage of our algorithm we use the detections that were selected in previous stages to restrict and enhance the segmentation map obtained from the appearance net. The effect of this step can be seen in figure~\ref{figure:filter}, in which the appearance map is filtered using the bounding boxes, and the background noise is removed.

To obtain the final prediction - a binary prediction segmentation mask - we threshold the appearance segmentation map twice, with a low and high threshold. Then we divide each of the obtained masks to their connected components. 

As a first pass, we take the high threshold mask and remove all the components that do not intersect with the bounding boxes that were selected in previous stages. This restriction should filter out erroneous segmented instances that are similar to the annotated object or simply noise. 

As a second pass, we add to the final segmentation mask the connected components from the low threshold mask that intersect with the mask obtained in the first pass. This enhancement provides a more lenient thresholding inside the selected bounding boxes and was inspired by Canny edge detector, which finds strong and weak edges and only selects the weak edges if they connect with strong ones. In our case, we look for strong and weak (high and low confidence) segmented pixels that are restricted to the selected bounding box area of the segmentation map and only select the weak pixels if their connected component intersects with the strong ones.

\section{Training} \label{training}
In this work we only trained the appearance network. The training was very similar to the one used in OSVOS~\cite{Cae+17}. For the offline training, we use Stochastic Gradient Descent with momentum 0.9. The augmentations we perform on the data are mirroring, rotations and resizing.

One difference from the Tensorflow implementation of OSVOS is that we do not use deep supervision for training. In that implementation the authors connect each side-output of the OSVOS architecture to a cross-entropy segmentation loss function and allow gradients to flow from multiple end sources. We noticed that when we simplify our network to only use feature maps of depth 1 (instead of 16) for each side output, we get similar results to the original OSVOS regardless of the use of deep supervision, but our training speed improves by an order of magnitude (see table~\ref{table:2}).

For the online one-shot training, we mostly keep the same parameters and enhance the training with online augmentations of the same kind as in the offline training. We have learned that it is best to keep the same local learning rates (i.e. the individual learning rate multipliers for each layer in the net) consistent between the online and offline training.

For the semantic object detection, we used the Tensorflow implementation of Faster-RCNN described in~\cite{tf-faster-rcnn}. Note: both the appearance network and semantic object detection network are modular and can be replaced with different implementations, as performance of their respective tasks improves in the future.
\section{Experiments and Results}
We conducted experiments on the DAVIS 2016 and DAVIS 2017~\cite{Pont-Tuset_arXiv_2017} datasets. While our ablation study was conducted only on DAVIS 2016 dataset, the conclusions should apply to DAVIS 2017 as well. 
\begin{figure}
\centering
\includegraphics[width=0.40\textwidth]{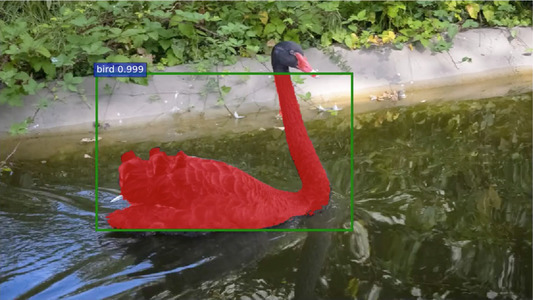}\\
\vspace{0.1cm}
\includegraphics[width=0.40\textwidth]{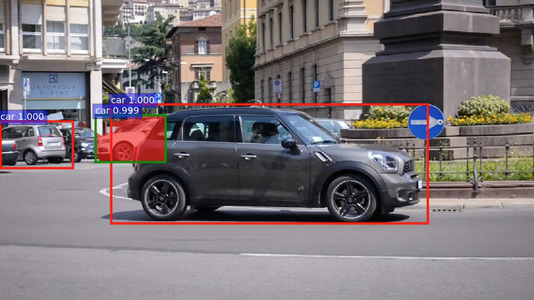} 
\medskip
\caption{Ablation examples. {\bf Top}: A decapitated black-swan swims in the lake when performing bounding box clipping. {\bf Bottom}: The annotated car is the one in the center, bounded by a red box. Unfortunately the car in the green box has a better IoU score between its bounding box and the appearance map. This type of failure can prevented by forcing temporal consistency on the bounding  box.}
\label{figure:ablation}
\end{figure}
In table~\ref{table:1} we summarize the experiments conducted to evaluate the various components of our methods. {\bf Appearance map} refers to solely using the appearance network trained on the one-shot annotated frame. Our appearance net builds on top of the official OSVOS-Tensorflow implementation, which gets $J_{mean}=75.8$ on DAVIS-2016 when we run it out-of-the-box. By streamlining the network as described in section \ref{Appearance Network} and changing the augmentations as described in \ref{training}, this benchmark was improved to 76.7. Running the same streamlined net on Caffe yielded 78 but we preferred to continue our experiments in the Tensorflow environment. {\bf BBox clipping} refers to limiting the segmentation of the appearance network by removing parts of the appearance map that lie outside the selected bounding box. {\bf Temporal} refers to adding the temporal bounding box selection method as explained in section~\ref{temporal}. {\bf ConnComp filter} refers to filtering components of the appearance segmentation map which do not intersect with the selected bounding boxes. 

Results from table~\ref{table:1} show that using the full method we reach a higher score than each component separately and that our temporal and connected component filters improve the score over using a simple bounding box clipping method. 

On the DAVIS-2016 validation dataset our whole system achieves a top score of $J_{mean}=80.1$ while on the DAVIS-2017 challenge dataset we reach $J_{mean}=49.7$. We have also built our own test set (see section ~\ref{app}) and evaluated it in the  manner of the DAVIS datasets to the score of $J_{mean}=86.6$.

In all the described experiments we have initialized our models with weights pre-trained on the ImageNet dataset and trained on DAVIS-2016 or DAVIS-2017 as described in section~\ref{training}. We did not exploit additional datasets for training such as COCO, PASCAL-VOC, or our own internal dataset. We did not apply further post-processing steps such as CRF.

Another advantage of our method is that it can be trained with less iterations than the baseline OSVOS~\cite{Cae+17} method. We are able to reduce the number of training iterations and the one-shot fine-tune iterations by limiting the depth of the side outputs to $1$ instead of $16$, and retaining only one loss layer (no deep supervision). The results in table~\ref{table:2} show a reduction of the training time by a factor of $10$. 

{\bf Running time:}
Since we can run both appearance and detection nets in parallel, we are as fast as the slowest network, which is the appearance network. The post-processing and bounding-box-tracking have a negligible running time, So the final run time for an average DAVIS video is very similar to OSVOS~\cite{Cae+17} and highly dependant on the number of online training iterations, resulting in an average runtime of between 1 and 10 seconds per frame. 

\begin{center}
\begin{table}
  \begin{tabular}{| l | c | c | c | c | c | c |}
    \hline \hline
    Appearance map & \checkmark & \checkmark & \checkmark & \checkmark & \checkmark \\ \hline 
    BBox clipping & \checkmark & \checkmark &   &  & \\ \hline
    Temporal &  & \checkmark & & & \checkmark \\ \hline 
    ConnComp filter &  &  & & \checkmark & \checkmark \\ \hline 
    $J_{mean}$  & 76.0 & 76.4 & 76.7 & 79.1 & 80.1 \\ 
    \hline \hline
  \end{tabular} \\
	\caption{Ablation study on the DAVIS 2016 dataset}
	\label{table:1}
\end{table}
\end{center}
\begin{table}
\centering
\begin{tabular}{| l | c | c |}
\hline \hline
& OSVOS & Our\\ \hline
iters. to $J_{mean}>79$ & 50K & 4K\\
\hline \hline
\end{tabular}\\
\medskip
\caption{Total number of training iterations until convergence}
\label{table:2}
\end{table}

\section{Building a Demonstration App and Evaluating Video Object Segmentation in the Wild} \label{app}
In order to test our algorithms in the real world, our team has built an internal mobile demonstration application that captures and processes videos. It allows the user to capture videos under 10 seconds in length, performs stabilization on the video which is centered around the object of interest, allows the user to provide a ground truth segmentation for the first frame and finally performs server-side object segmentation for the entire video and presents the results with various effects.

Through our work with and on the app we have learned about the viability of the segmentation algorithms in natural user-captured videos. First, we have built our own internal dataset of about 150 user captured videos, complete with human-traced ground truth annotations. On the one hand, most of our videos are simpler than in DAVIS - virtually devoid of occlusions, fast motion or similar repeating instances of the same object type. On the other hand, the objects in our videos have more varied categories than the DAVIS-2016 dataset, in which many of the videos contain known semantic classes (humans, cars, etc). We have tested the method presented in this paper on this new dataset and achieved a result of $J_{mean}=86.6$, which is not far from what was achieved on DAVIS-2016 (table~\ref{table:1}). To corroborate our results with external research, {\bf we plan to publish our internal dataset to a public repository in the near future} for all to use freely.

Second, we have tested the hypothesis of semi-supervised video segmentation with actual users and we believe there is real potential for it. After capturing a video, the user is given a simple interface to mark the object in a semi-automatic way. We have implemented a variant of the grab-cut algorithm and demonstrated a few quick scribbles on the smartphone screen enable a user to efficiently annotate the first frame of a captured video. We have been positively impressed by the quality of segmentations achieved by this method in mere seconds. Another pleasant surprise was that the video object segmentation continued to be of mostly high quality even when the first frame segmentation was not completely accurate, showing some robustness.


\section{Conclusions \& future work}
We presented a method for category-independent object segmentation in videos which combines the outputs of a one-shot trained appearance net and a semantic instance detection net, while imposing temporal constraints on the results. We have demonstrated the benefit of each of the building blocks and also suggested a way to drastically increase the training speed for the appearance net. 

In our future work, we plan to explore a multi-task network architecture that can be trained end-to-end in a one-shot fashion. This architecture merges the object detection and the appearance networks into a single joint network with two heads, which would allow the detection part of the network to fine-tune on the annotated object in the first frame and thus improve the detection and in turn, the segmentation accuracy.





{\small
\bibliographystyle{ieee}
\bibliography{davis2017}
}

\end{document}